%% file: main.tex
\documentclass[final]{cvpr}

\usepackage{times}
\usepackage{epsfig}
\usepackage{graphicx}
\usepackage{amsmath}
\usepackage{amssymb}
\usepackage{kotex}
\usepackage{multirow}
\usepackage{comment}
\usepackage{subcaption}
\usepackage{booktabs}

\usepackage[pagebackref=true,breaklinks=true,colorlinks,bookmarks=false]{hyperref}

\begin{document}

\title{On-Off Pattern Encoding and Path-Count Encoding\\as Deep Neural Network Representations\thanks{This research was conducted in 2020, and the content remains unchanged from the original manuscript.}}

\author{
Euna Jung\textsuperscript{\rm 1},
Jaekeol Choi\textsuperscript{\rm 1,2},
EungGu Yun\textsuperscript{\rm 3}\thanks{Work performed during the internship program at Seoul National University.},
Wonjong Rhee\textsuperscript{\rm 1,4}\thanks{Corresponding author}\\
\textsuperscript{\rm 1} GSCST,
Seoul National University\\
\textsuperscript{\rm 2} Naver Corporation \\
\textsuperscript{\rm 3} Kim Jaechul Graduate School of AI, KAIST \\
\textsuperscript{\rm 4} IPAI, Seoul National University \\
{\tt\small xlpczv@snu.ac.kr},
{\tt\small jaekeol.choi@snu.ac.kr},
{\tt\small eunggu.yun@kaist.ac.kr}, 
{\tt\small wrhee@snu.ac.kr}
}

\maketitle

\input{0_Abstract}
\input{1_Introduction}
\input{2_Related_Works}
\input{3_Methods}
\input{4_Experiment}
\input{5_Conclusion}

\newpage
{\small
\bibliographystyle{ieee_fullname}
\bibliography{main}
}

\end{document}

%% file: 0_Abstract.tex
\begin{abstract}
Understanding the encoded representation of Deep Neural Networks (DNNs) has been a fundamental yet challenging objective.
In this work, we focus on two possible directions for analyzing representations of DNNs by studying simple image classification tasks. Specifically, we consider \textit{On-Off pattern} and \textit{PathCount} for investigating how information is stored in deep representations.
On-off pattern of a neuron is decided as `on' or `off' depending on whether the neuron's activation after ReLU is non-zero or zero. PathCount is the number of paths that transmit non-zero energy from the input to a neuron. We investigate how neurons in the network encodes information by replacing each layer's activation with On-Off pattern or PathCount and evaluating its effect on classification performance. We also examine correlation between representation and PathCount. Finally, we show a possible way to improve an existing DNN interpretation method, Class Activation Map (CAM), by directly utilizing On-Off or PathCount.
\end{abstract}

%% file: 1_Introduction.tex
\section{Introduction}
\begin{figure*}[!t]
\begin{center}
   \includegraphics[width=0.9\linewidth]{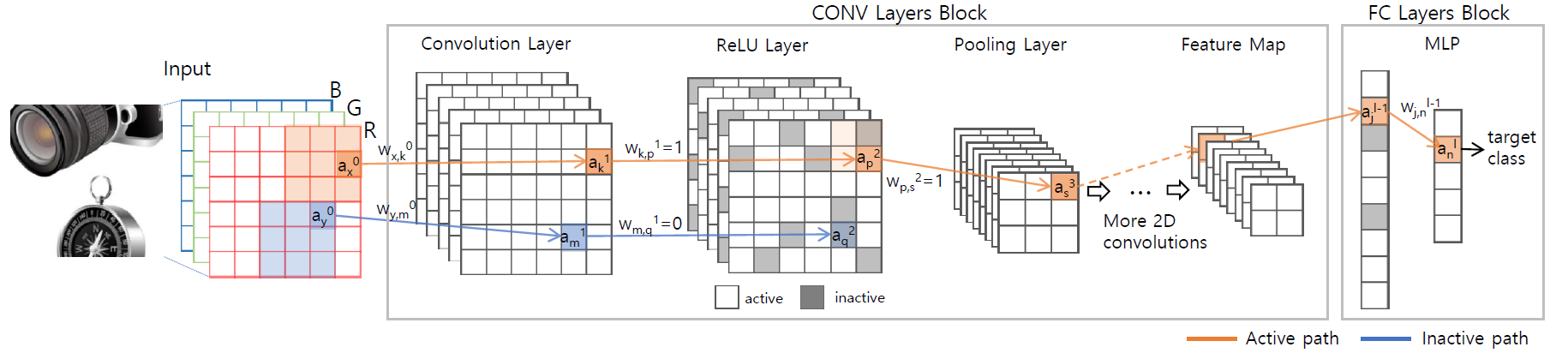}
\end{center}
   \caption{Definition of a path. We define a path as a sequence of neurons between two neurons. In the figure, we show an example of an active path from one input element and one neuron at the last layer with an orange line. A blue line represents an inactive path that is not propagated through the network because one of the activations consisting of a path sequence is zero.}
\label{fig: def. of a path}
\end{figure*}
The utilization of Deep Learning (DL) has become widespread across numerous fields. In particular, DNNs for vision tasks such as autonomous driving and face recognition have been intensively studied. As DNNs for these tasks are closely related to security and safety, many studies have tried to understand DNNs. Despite the effort, DNN is still considered as a black-box algorithm. In this paper, we investigate two possible directions for understanding how information is stored in a neuron's activation. 

In DNNs, information needed for the task of interest is encoded in the representation, i.e., neuron values of each layer. We call representation of the layer before the activation function as \textit{pre-activation} and output values of the activation function as \textit{activation}. DNN's $l^{th}$ layer activation is formed by applying the activation function to the $l^{th}$ layer pre-activation, which is computed by affine transformation of ${(l-1)}^{th}$ layer activation. In short, DNN's representation is formulated by affine transformation and activation function, and we analyze how representation encodes information.

Activation function is important for deep learning because activation function enables DNNs to model non-linearity \cite{sharma2017activation, nwankpa2018activation}. There are frequently used activation functions for DNNs such as hyperbolic tangent, sigmoid, and ReLU \cite{nair2010rectified} functions. Among them, ReLU is the most successful and widely used function in that it offers better performance and generalization compared to other functions \cite{ramachandran2017searching,zeiler2013rectified,dahl2013improving,nwankpa2018activation}. In fact, benchmark models for image classification such as AlexNet \cite{krizhevsky2012imagenet}, VGG \cite{simonyan2014very}, ResNet \cite{he2016deep}, GoogLeNet \cite{szegedy2015going}, and recent MobileNet \cite{howard2017mobilenets} all use ReLU as an activation function. Therefore, we analyze how activation of ReLU output encodes information.

ReLU is a function that decides if a neuron can be fired or not \cite{nwankpa2018activation}, and only fired neurons can propagate information forward and backward. We focus on this characteristic of ReLU function, analyzing the role of \textit{On-Off} pattern in DNNs. On-Off pattern of a neuron is decided as zero/one if a neuron value is zero/positive. In other words, On-Off represents whether the neuron can propagate the information or not in a binary form. By replacing activation with On-Off, we examine how much DNNs rely on ReLU On-Off when encoding information.

Furthermore, in DNNs using convolution and ReLU, each neuron has different receptive fields, and ReLU makes a hard decision on ignition of each neuron. Therefore, each neuron is connected from the input with different number of paths. We define \textit{PathCount} as the number of active paths that transmit non-zero signals from one neuron to the other. We measure PathCount for each neuron as the number of paths from the input to the neuron and analyze how much DNN encoding relies on PathCount as well as On-Off pattern.

Utilizing On-Off and PathCount, we devise three experiments to understand DNN representation. First, by replacing layerwise activation with On-Off and PathCount, we examine how neurons encode information of input for the task. When the performance of a network maintains high after replacement, we can infer that activation depends on that replacement for information encoding. Second, we further analyze the relationship between representation and PathCount using Kendall's Tau correlation test. If the correlation is high, we can say that PathCount contributes a lot to the formation of DNN representation. Finally, we suggest modified Class Activation Maps (CAMs) applying On-Off and PathCount. By changing the level of feature extraction layer and evaluating the performance of modified methods, we confirm the role of On-Off and PathCount again, proving usefulness of our methodologies.

Experimental results show that DNN encoding heavily relies on On-Off and PathCount in the upper layers. Even in some models, inference accuracy is preserved after replacing activation with On-Off or PathCount. This result is supported by the experiment using modified CAMs.

Our contributions can be summarized as follows:
\begin{itemize}
    \item[1] We suggest new methodologies for analyzing DNNs. Especially, we provide an efficient method to count all paths from two neurons, enabling microscopic perspective on neural networks.
    \item[2] We analyze how neurons encode information needed for the task. We figure out On-Off binary pattern and PathCount of a neuron is importantly used for encoding.
    \item[3] We examine the relationship between representation and PathCount, by which we provide a clue for understanding the formation of representation in DNNs.
    \item[4] We suggest a new method for interpreting neural networks by modifying CAM. Our modified CAMs show similar or even better performance while using less information in activation.
\end{itemize}
In Section 2, we summarize related works. In Section 3, we explain how to define On-off pattern and PathCount in detail. In Section 4, we present three experiments and results to figure out how representation encode the information. Finally we conclude and discuss in Section 5.

%% file: 2_Related_Works.tex
\section{Related Works}
We introduce studies related to our work. We first suggest papers studying representation of different layers and models. Next, we present a few works related to On-Off pattern and paths that are methodologies used in this paper.

\subsection{Works for understanding deep neural networks}
Many studies have focused on interpreting DNNs. In particular, methodologies for analyzing and comparing how layerwise features of different models encode information have been developed. Zeiler and Fergus \cite{zeiler2014visualizing} and Erhan \etal \cite{erhan2009visualizing} figured out that each layer in deep convolutional networks extracts different levels of features using deconvolution and activation maximization methods, respectively. They showed that lower layers of convolutional networks extract simple features such as edges while upper layers combine simple features to model complicated concepts.

Other studies compared representations learned by different DNNs. Yu \etal \cite{yu2014visualizing} compared VGG and Alexnet by probing internal representation of each network. McNeely-White \etal \cite{mcneely2020inception} revealed that ResNet and Inception extract almost equivalent features using feature mapping. Li \etal \cite{li2017does} compared VGG, GoogLeNet, and ResNet, concluding that features learned by ResNet were less applicable to other vision tasks without fine-tuning compared to other two models. In this paper, we compare how layerwise representation encodes information and identify encoding characteristics of different models for different datasets using On-Off and PathCount.

\subsection{Works on On-Off pattern}
There are studies dealing with On-off patterns of DNNs focusing the number of linear regions. Pascanu and Montufar \etal \cite{pascanu2013number, montufar2014number} defined a \textit{linear region} of a piecewise linear function as a ``maximal connected subset of the input-space'' and counted the number of linear regions. In other words, each linear region of a rectifier network corresponds to an On-off pattern \cite{montufar2014number}. Therefore, the number of linear regions equals to the number of possible On-off patterns for the given input space and weights.
A series of studies \cite{pascanu2013number, montufar2014number, hanin2019deep, hanin2019complexity} have confirmed that the practical number of linear regions of DNNs are far fewer than the theoretical number. From this result, we can assume that On-off patterns of a network contain wealthy information for classification. We provide results supporting this assumption in this paper.

\subsection{Works on paths}
The concept of \textit{paths} of a neural network has not been pervasively used in the studies so far, but a few papers dealt with it. Wang \etal \cite{wang2018interpret} defined \textit{critical data routing paths} using critical nodes that were composed of important channels of the convolutional neural network. The concept of paths of this study is different from the paths suggested in our study, but they both focus on the connections in the neural network. Yuxian \etal \cite{qiu2019adversarial} following \cite{wang2018interpret} defines \textit{effective paths} using critical neurons and weights. Inspired by the idea of selecting critical neurons and weights, we clip weights based on their absolute values when counting the number of paths through FC layers. Our study is the first to count the number of paths in the neural network.

%% file: 3_Methods.tex
\section{Methodology}
\begin{figure*}
\centering
\begin{tabular}{cccc}
    \includegraphics[width=0.3\linewidth]{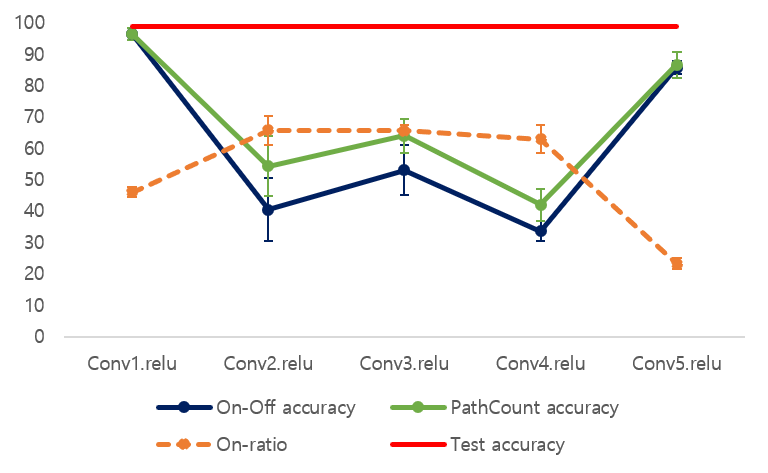}&
    \includegraphics[width=0.3\linewidth]{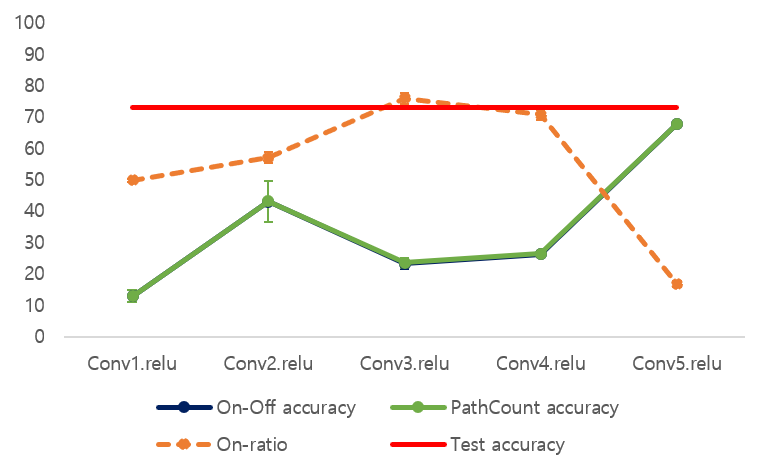}&
\\
(a) CONV-FC on MNIST & (b) CONV-FC on CIFAR10
\end{tabular}
\\
\begin{tabular}{cccc}
    \includegraphics[width=0.3\linewidth]{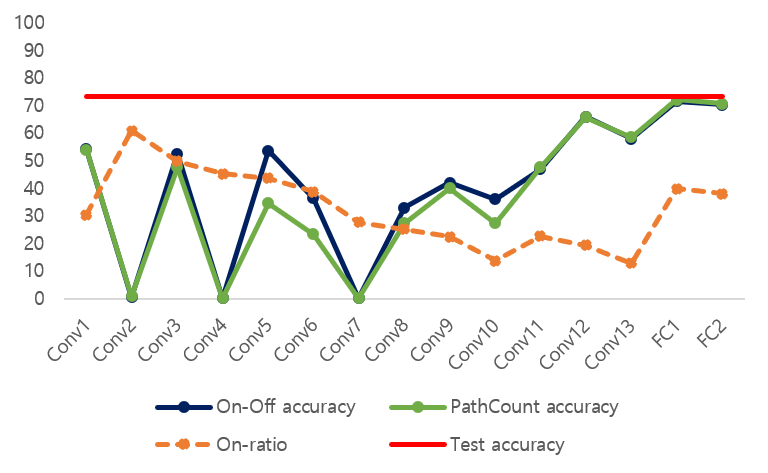}&
    \includegraphics[width=0.3\linewidth]{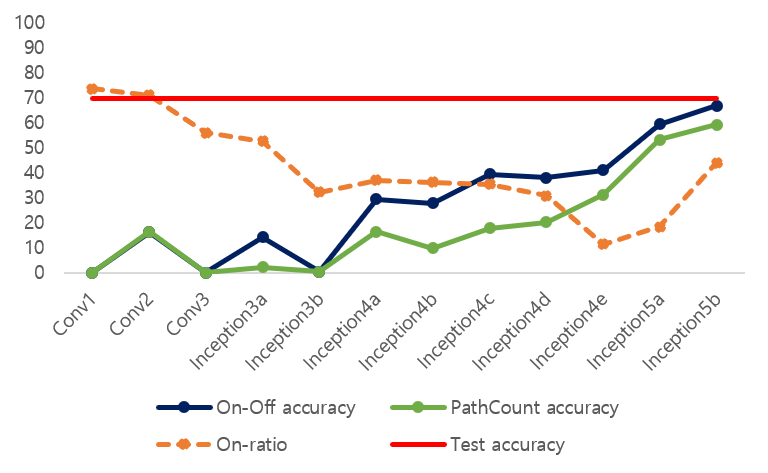}&
    \includegraphics[width=0.3\linewidth]{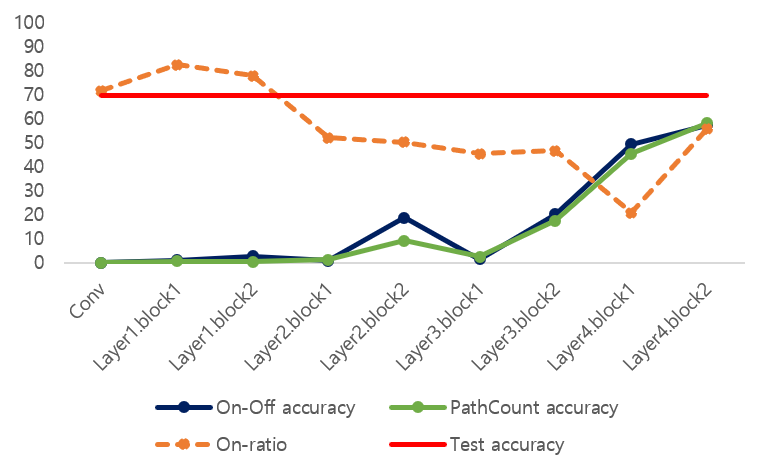}
\\
(c) VGG16 on ImageNet & (d) GoogLeNet on ImageNet & (e) ResNet18 on ImageNet
\end{tabular}
\newline
\caption{Classification accuracy when replacing layerwise activation with On-Off and PathCount. Image classification accuracy after the replacement of layerwise activation with On-Off and PathCount is presented. For analysis, we also present On-ratio, which means the ratio of neurons having positive values over the number of neurons in each layer.}
\label{fig: On-Off accuracy and On-ratio}
\end{figure*}

We assume that information for the tasks is encoded in the representation of DNNs. To figure out how representation is encoded, we define On-Off patterns and PathCount, with which we devise experiments. As activation functions are used in neural networks to decide if a neuron can be fired or not \cite{nwankpa2018activation}, and ReLU that makes hard decision on the neurons' ignition are used for almost all benchmark models, we come up with two methods that include the concept of On-off patterns.

\subsection{On-Off pattern}
On-Off pattern of a neuron $a_{i}^{l}$ in the pre-trained DNN $o(a_{i}^{l})$ is defined as:
\begin{equation}
  o(a_{i}^{l}) = \begin{cases}
    1, & \text{if $a_{i}^{l}>0$}.\\
    0, & \text{otherwise}.
  \end{cases}
  \label{eq:OnOff}
\end{equation}
where $a_{i}^{l}$ is $i^{\text{th}}$ activation in layer $l$. $a_{i}^{0}$ refers to $i^{\text{th}}$ input element. On-off pattern of a neuron represents whether the neuron can propagate the information or not. If the output of ReLU activation function is zero, then that neuron cannot transmit the information either forward or backward. We assume that DNNs contain a large amount of information in the representation using binary coding of On-Off patterns, so that we figure out how DNNs react when we replace activation with On-off patterns.

\subsection{PathCount}
For a neural network, we define a path as a sequence of neurons between two neurons that are located in two different layers. As shown in Figure \ref{fig: def. of a path}, only one neuron from each layer can be selected to form a path, and the selected neurons in layers $l$ and $l+ 1$ should be connected through a neural network operation such as convolution.
Paths between $a_{i}^{l}$ and $a_{j}^{l+m}$, $p$($a_{i}^{l}$, $a_{j}^{l+m}$), are written:
\begin{multline}
    p(a_{i}^{l}, a_{j}^{l+m}) \\
    = \{ \: (a_{i}^{l}, a_{k}^{l+1}, ..., a_{q}^{l+m-1}, a_{j}^{l+m}) \: | \: k, \cdots, q \in \mathbb{Z_{+}} \} \\
  \label{eq: Paths between two neurons}
\end{multline}
For convenience, we define $p(a_{j}^{l})$ as the set of all paths from all input elements to a neuron $a_{j}^{l}$:
\begin{multline}
    p(a_{j}^{l}) = \{ \: (a_{i}^{0}, a_{k}^{1}, ..., a_{q}^{l-1}, a_{j}^{l}) \: | \: i, k, \cdots, q \in \mathbb{Z_{+}} \} \\
  \label{eq: Paths for a neuron}
\end{multline}

If a path does not transmit energy during propagation because a neuron in a path becomes zero not surviving ReLU function, or a weight between neurons is zero, that path is called \textit{inactive}. On the other hand, if a path is composed of nonzero neurons all connected by nonzero weights, that path is called \textit{active}. If a neuron is connected by many active paths from the input, more paths are contributing to the value of that neuron, so that we define PathCount using only active paths and analyze the network with it.
PathCount of $a_{j}^{l}$, $pc(a_{j}^{l})$, is defined as the number of active paths from all input elements to a neuron $a_{j}^{l}$:
\begin{multline}
    pc(a_{j}^{l}) = |\{ \: (a_{i}^{0}, a_{k}^{1}, ..., a_{j}^{l}) \: | \: a_{k}^{l} \neq 0 \: \text{and} \: w(a_{k}^{l}, a_{q}^{l+1}) \neq 0, \\
    i, k \cdots, q \in \mathbb{Z_{+}} \}|
  \label{eq: PathCount of a neuron}
\end{multline}
where $w(a_{k}^{l}, a_{q}^{l+1})$ is the weight connecting two neurons $a_{k}^{l}$ and $a_{q}^{l+1}$. On-Off pattern of a neuron in a path can decide whether that path is active or inactive, so that PathCount is a concept that includes On-Off pattern in it.

We develop a method to count paths between two neurons very efficiently. In simple words, we can count how many active paths exist between two neurons by changing weights into a constant. After changing all weights into ones, we can easily count the number of paths from each input element to a neuron by computing the gradients of a neuron by the input. If we forward propagate the input whose elements are all `ones' to a certain layer, we can efficiently calculate PathCount from the input to the neurons in the layer.

As convolutions (CONV) generate different receptive fields for different neurons, each convolution output has different PathCount. On the other hand, output neurons of fully connected (FC) layer are connected with all input neurons of FC since neurons through FC layer all have the same PathCount. To cope with this problem, we compute PathCount of FC output layers after clipping FC weights. We suppose that two layers are connected through FC by weights bigger than a certain threshold.

As with On-Off pattern, we replace activation with PathCount to figure out how activation depends on PathCount when encoding the information in Section 4. Furthermore, we conduct Kendall's Tau test on PathCount and the activation to investigate the relationship between them.

%% file: 4_Experiment.tex
\section{Experiment}
\begin{table*}[!t]
  \begin{subtable}{.5\linewidth}
    \centering
    \begin{tabular}{l|rr}
      \toprule
      \multirow{2}{*}{Module}& \multicolumn{2}{c}{Kendall's Tau} \\ \cline{2-3}
      & \multicolumn{1}{c}{Raw} & \multicolumn{1}{c}{Abs} \\ \midrule \midrule
        Conv1.conv             & -0.010±0.012  & -0.080±0.002 \\
        Conv1.relu             & 0.806±0.010   & 0.806±0.010  \\ \midrule
        Conv2.conv             & 0.105±0.019  & 0.400±0.112   \\
        Conv2.relu             & 0.719±0.068  & 0.719±0.068 \\ \midrule
        Conv3.conv             & 0.056±0.054  & 0.065±0.221 \\
        Conv3.relu             & 0.536±0.132  & 0.536±0.132 \\ \midrule
        Conv4.conv             & 0.087±0.060   & 0.067±0.240  \\
        Conv4.relu             & 0.578±0.116  & 0.578±0.116 \\ \midrule
        Conv5.conv             & -0.048±0.084 & 0.046±0.142 \\
        Conv5.relu             & 0.874±0.013  & 0.874±0.013 \\
      \bottomrule
    \end{tabular}
    \caption{MNIST}
    \label{tab: Kendall Tau - MNIST}
  \end{subtable}%
  \begin{subtable}{.5\linewidth}
    \centering
    \begin{tabular}{l|rr}
      \toprule
      \multirow{2}{*}{Module}& \multicolumn{2}{c}{Kendall's Tau} \\ \cline{2-3}
      & \multicolumn{1}{c}{Raw} & \multicolumn{1}{c}{Abs} \\ \midrule \midrule
        Conv1.conv               & -0.000±0.000     & -0.032±0.001 \\
        Conv1.relu               & 0.769±0.003  & 0.769±0.003  \\ \midrule
        Conv2.conv               & -0.009±0.007 & 0.001±0.013  \\
        Conv2.relu               & 0.599±0.019  & 0.599±0.019  \\ \midrule
        Conv3.conv               & 0.020±0.012   & 0.013±0.014  \\
        Conv3.relu               & 0.388±0.015  & 0.388±0.015  \\ \midrule
        Conv4.conv               & 0.028±0.005  & 0.016±0.009  \\
        Conv4.relu               & 0.454±0.020   & 0.454±0.020   \\ \midrule
        Conv5.conv               & -0.002±0.009 & -0.002±0.011 \\
        Conv5.relu               & 0.904±0.008  & 0.904±0.008  \\
      \bottomrule
    \end{tabular}
    \caption{CIFAR10}
    \label{tab: Kendall Tau - CIFAR10}
  \end{subtable}
  \caption{Correlation between representation and PathCount on MNIST and CIFAR10.}
  \label{tab: Kendall Tau - MNIST, CIFAR10}
\end{table*}

\begin{table}[ht]
\centering
\footnotesize
\begin{tabular}{l|rr}
\toprule
\multirow{2}{*}{Module}& \multicolumn{2}{c}{Kendall's Tau} \\ \cline{2-3}
& Raw             & Abs             \\ \midrule  \midrule
Conv1.conv                  & 0.000         & -0.004      \\
Conv1.batchnorm             & 0.003       & -0.003      \\
Conv1.relu                  & 0.907       & 0.907       \\ \midrule
Conv2.conv                  & 0.010        & -0.027      \\
Conv2.batchnorm             & 0.024       & -0.049      \\
Conv2.relu                  & 0.576       & 0.576       \\ \midrule
Conv3.conv                  & 0.007       & -0.019      \\
Conv3.batchnorm             & 0.023       & -0.037      \\
Conv3.relu                  & 0.685       & 0.685       \\ \midrule
Conv4.conv                  & 0.043       & -0.076      \\
Conv4.batchnorm             & 0.050        & -0.111      \\
Conv4.relu                  & 0.701       & 0.701       \\ \midrule
Conv5.conv                  & 0.029       & -0.072      \\
Conv5.batchnorm             & 0.034       & -0.079      \\
Conv5.relu                  & 0.720        & 0.720        \\ \midrule
Conv6.conv                  & -0.002      & -0.018      \\
Conv6.batchnorm             & -0.008      & -0.051      \\
Conv6.relu                  & 0.739       & 0.739       \\ \midrule
Conv7.conv                  & 0.023       & -0.042      \\
Conv7.batchnorm             & 0.024       & -0.066      \\
Conv7.relu                  & 0.822       & 0.822       \\ \midrule
Conv8.conv                  & 0.007       & -0.051      \\
Conv8.batchnorm             & 0.007       & -0.042      \\
Conv8.relu                  & 0.848       & 0.848       \\ \midrule
Conv9.conv                  & 0.003       & -0.017      \\
Conv9.batchnorm             & 0.002       & -0.028      \\
Conv9.relu                  & 0.866       & 0.866       \\ \midrule
Conv10.conv                 & -0.015      & 0.013       \\
Conv10.batchnorm            & -0.016      & -0.007      \\
Conv10.relu                 & 0.918       & 0.918       \\ \midrule
Conv11.conv                 & -0.054      & 0.030        \\
Conv11.batchnorm            & -0.053      & 0.034       \\
Conv11.relu                 & 0.866       & 0.866       \\ \midrule
Conv12.conv                 & -0.069      & 0.070        \\
Conv12.batchnorm            & -0.071      & 0.060        \\
Conv12.relu                 & 0.895       & 0.895       \\ \midrule
Conv13.conv                 & -0.063      & 0.091       \\
Conv13.batchnorm            & -0.066      & 0.113       \\
Conv13.relu                 & 0.923       & 0.923       \\ \midrule
FC1                         & 0.882       & 0.882       \\
FC2                         & 0.906       & 0.906       \\
\bottomrule
\end{tabular}
\newline
\caption{Correlation between layerwise representation and PathCount of VGG16 on ImageNet. Each CONV layer of VGG16 is composed of three operations, convolution, batch normalization, and ReLU. We compare representation and PathCount for every operation. FC layers of VGG16 consist of a fully connected layer and ReLU. For FC layers, we show correlation after ReLU.}
\label{tab: ImageNet-VGG16}
\end{table}

\begin{table}[ht]
\centering
\small
\begin{tabular}{l|rr}
\toprule
\multirow{2}{*}{Module}& \multicolumn{2}{c}{Kendall's Tau} \\ \cline{2-3}
& Raw             & Abs             \\ \midrule  \midrule
Conv1.conv      & 0.001  & -0.004 \\
Conv1.batchnorm & 0.002  & -0.007 \\
Conv1.relu      & 0.608  & 0.608  \\ \midrule
Conv2.conv      & -0.031 & 0.041  \\
Conv2.batchnorm & -0.044 & 0.056  \\
Conv2.relu      & 0.551  & 0.551  \\ \midrule
Conv3.conv      & 0.070  & -0.058 \\
Conv3.batchnorm & 0.084  & -0.116 \\
Conv3.relu      & 0.615  & 0.615  \\ \midrule
Inception3a     & 0.640  & 0.640  \\
Inception3b     & 0.794  & 0.794  \\ \midrule
Inception4a     & 0.764  & 0.764  \\
Inception4b     & 0.765  & 0.765  \\
Inception4c     & 0.769  & 0.769  \\
Inception4d     & 0.809  & 0.809  \\
Inception4e     & 0.940  & 0.940  \\ \midrule
Inception5a     & 0.904  & 0.904  \\
Inception5b     & 0.781  & 0.781  \\
\bottomrule
\end{tabular}
\newline
\caption{Correlation between layerwise representation and PathCount of GoogLeNet on ImageNet. Each CONV layer of GoogLeNet is composed of three operations: convolution, batch normalization, and ReLU. We compare representation and PathCount for every operation. Additionally, we show the results for representations after each inception block.}
\label{tab: ImageNet-GoogleNet}
\end{table}

We devise three experiments to figure out how DNN representation encodes information using On-Off pattern and PathCount suggested in Section 3. To clearly identify the effect of PathCount on representation values, we first experiment with pre-trained networks without bias or batch normalization on MNIST and CIFAR10 datasets. Next, we expand our experiments to benchmark models on the ImageNet (ILSVRC-2012) data. For these complicated models, we use some tricks to appropriately compute the PathCount.

\subsection{Experimental details}
Throughout the experiment section, we show the results of five pairs of benchmark datasets and models (CONV-FC on MNIST/CIFAR10 and VGG16/GoogLeNet/ResNet18 on ImageNet).

CONV-FC model consists of five CONV layers and one FC layer. For accurate computation of PathCount, we employed convolutional operations without bias and batch normalization. At the last CONV output layer, maxpooling, dropout with a droprate of 0.25 is attached. Kernel size of 3 and stride 1 are used for all CONV layers, and 32/64 channels are used for CONV layers for MNIST/CIFAR10. The models were trained with the learning rate of 0.001 using stochastic gradient descent for 100 epochs. We trained the model with 50,000 train data and evaluated accuracy with 10,000 test data. For calculating Kendall’s Tau, we used randomly selected 1,000 test data. Because CONV-FC models are trained three times with different seeds, we present the mean and standard deviation of Tau.

We use pre-trained networks provided by Torchvision for ImageNet dataset.

\subsection{Activation replacement with On-Off patterns and PathCount}
To figure out how layerwise activation encodes information, we devise an experiment of replacing activation with On-Off pattern and PathCount in each layer of pre-trained network and evaluating the inference accuracy. We call the accuracy after replacement as `On-Off accuracy' and `PathCount accuracy' respectively. Because we do not fine-tune the network after changing activation values, layerwise On-Off/PathCount accuracy expresses the lower bound for the information contained in On-Off/PathCount in each layer.
To adjust the scales of replaced representation, we use \textit{scaled On-Off} $\tilde{o}(a_{i}^{l})$ and \textit{scaled PathCount} $\tilde{pc}(a_{i}^{l})$ for the inference by multiplying a constant to On-Off $o(a_{i}^{l})$ and PathCount $pc(a_{i}^{l})$ in each layer:
\begin{equation}
\label{eq: scaled On-Off}
\tilde{o}(a_{i}^{l}) = o(a_{i}^{l}) \cdot \frac{\mu_{a}^{l}}{\mu_{o}^{l}}
\end{equation}
\begin{equation}
\label{eq: scaled PathCount}
\tilde{pc}(a_{i}^{l}) = pc(a_{i}^{l}) \cdot \frac{\mu_{a}^{l}}{\mu_{pc}^{l}}
\end{equation}
where $\; \mu_{o}^{l} = \sum_{i: a_{i}^{l} \neq 0} o(a_{i}^{l}), \; \mu_{pc}^{l} = \sum_{i} pc(a_{i}^{l})$.

In Figure~\ref{fig: On-Off accuracy and On-ratio}, we present On-Off accuracy and PathCount accuracy with the test accuracy that is the inference accuracy of the original pre-trained network. For comparison, we also provide `On-ratio' that equals to the ratio of nonzero neurons over the number of neurons in the ReLU output layer.

Figure~\ref{fig: On-Off accuracy and On-ratio} (a) shows the On-Off accuracy result of the model on MNIST.
We first observe that On-Off pattern embodies much information of the activation. Especially On-Off of the first and last layer have more than 90\% of information needed for image classification. Second, PathCount contains even more information of representation than On-Off. Third, when we preserve the sign of representation when replacing it with PathCount, the accuracy increases. These results are in accordance with the high correlation between absolute representation and PathCount.

Figure~\ref{fig: On-Off accuracy and On-ratio} (b) presents the On-Off accuracy result of the model on CIFAR10. As in Figure~\ref{fig: On-Off accuracy and On-ratio} (a), the accuracy on the last convolution output is close to the accuracy of the original network. The general shape of the graph resembles the graph on MNIST in that On-Off accuracy decreases in intermediate layers and soars on the top ReLU output layer while On-ratio moves in the opposite direction. However, the accuracy of PathCount is not better than that of On-Off, which is in accordance with the correlation result in Section 4.1.

In Figure~\ref{fig: On-Off accuracy and On-ratio} (c), we provide the activation replacement accuracy result of VGG16 on ImageNet. Even in a deep neural network (VGG16), for the complicated task (ImageNet classification), we can observe the tendency of the last layer to contain a lot of information using binary encoding. In this deep network, we can easily find fluctuation in the On-Off accuracy. In agreement with the correlation result, PathCount accuracy is not better than On-Off accuracy (even worse in some layers).

\begin{figure*}[!t]
\begin{center}
   \includegraphics[width=0.9\linewidth]{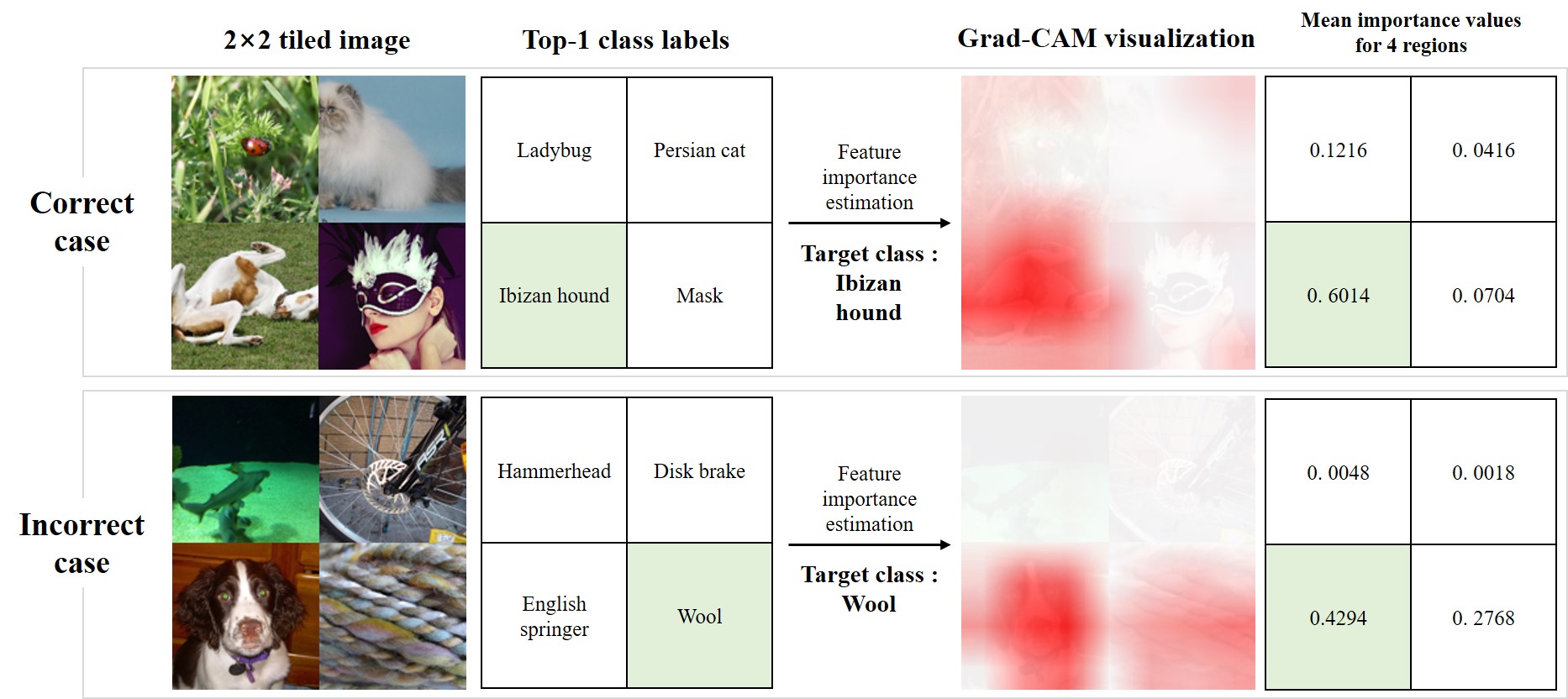}
\end{center}
\caption{An example of target-matching process on Tiled-ImageNet. The upper picture represents a correct case of target-matching on Tiled-ImageNet where the top-1 class label used for computing importance of each pixel matches the estimated target class based on Grad-CAM scores. The lower picture shows an incorrect case of target-matching where the estimated target class is `English springer' while the real target class was `Wool'. We evaluate the target-matching accuracy by computing the ratio of correct cases.}
\label{fig:tiled-imagenet target-matching accuracy}
\end{figure*}

\begin{figure}[ht]
\begin{center}
   \includegraphics[width=0.45\textwidth]{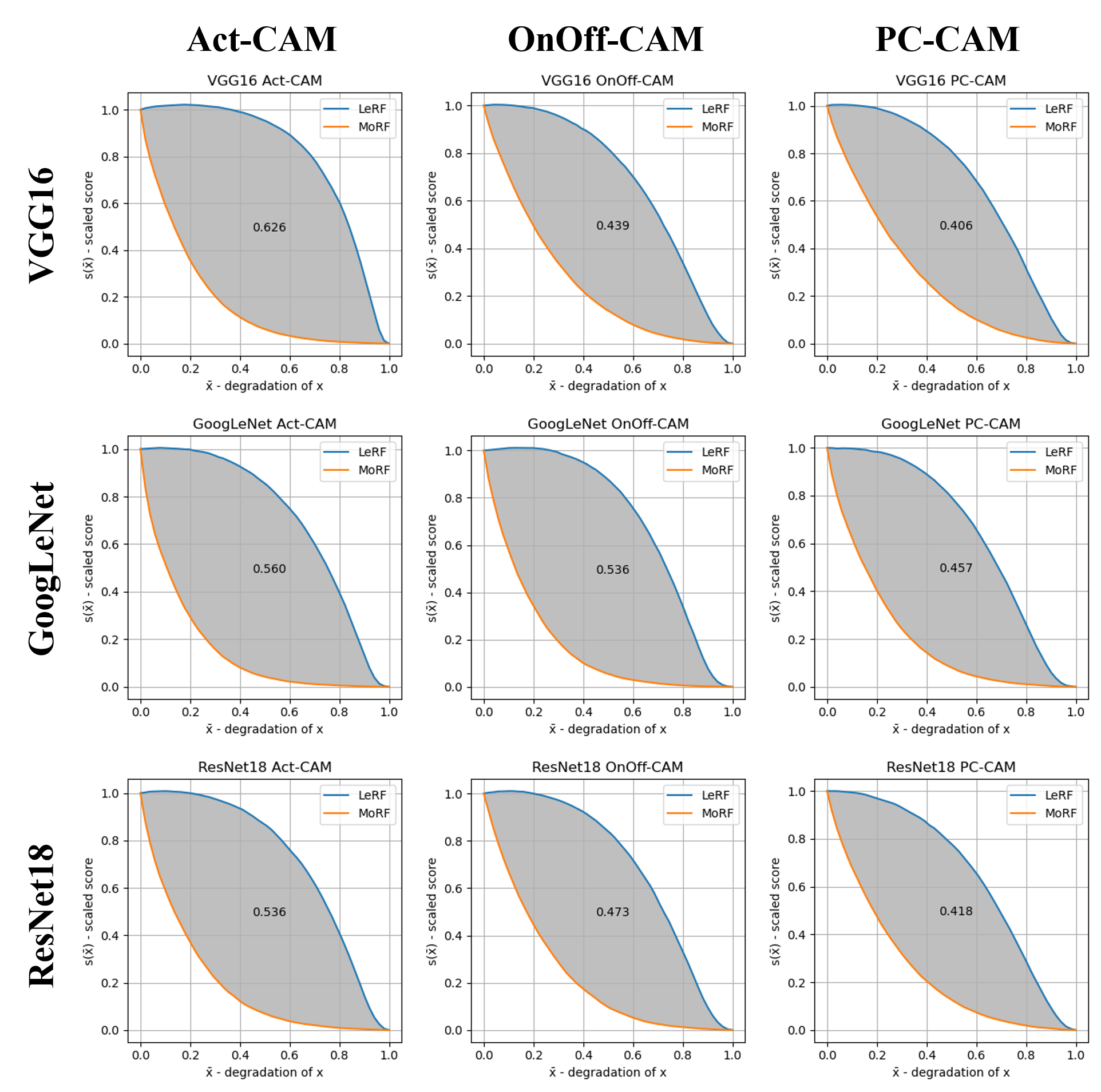}
\end{center}
\caption{Degradation performance of Act-CAM, OnOff-CAM, and PC-CAM. We present the degradation performance of three modified CAMs on three networks, VGG16, GoogLeNet, and ResNet18 trained on ImageNet. Act-CAM that uses activation of the last convolution output shows the best degradation performance followed by OnOff-CAM and PC-CAM.}
\label{fig:CAM}
\end{figure}

\subsection{Correlation between representation and PathCount}

We assume that with the more paths a neuron is connected from the input, the bigger the representation is.
To confirm this idea, we evaluate the correlation between the representation and PathCount using Kendall's Tau~\cite{abdi2007kendall}, a widely used metric for assessing the ordinal association between two quantities. Since Kendall's Tau calculates the correlation based on the rankings, we can effectively evaluate the relationship between representation and PathCount regardless of the scaling problem.

To check if PathCount affects only the magnitude of the representation or not, we compare both raw and absolute values of representation with PathCount. In Table~\ref{tab: Kendall Tau - MNIST}, we show that PathCount and absolute values of the representation are highly correlated, meaning a neuron connected with many paths from the input tends to have a bigger absolute value (We should ignore the first layer because PathCount to the first convolution layer is almost the same for all neurons). 

Different from the result on MNIST, Table~\ref{tab: Kendall Tau - CIFAR10} shows low correlation between representation and PathCount of the pre-trained model on CIFAR10. This means that representation is not proportionally decided by the number of active paths from the input in the pre-trained network on CIFAR10. Correlation for ReLU output layer is high because there are many ties of zero activations. 

We experimented with two benchmark models, VGG~\cite{simonyan2014very} and GoogLeNet~\cite{szegedy2015going} for ImageNet data classification and show the results in Table~\ref{tab: ImageNet-VGG16} and Table~\ref{tab: ImageNet-GoogleNet}.
As in the pre-trained model for CIFAR10, the correlation between representation and PathCount is close to zero in the network for ImageNet.

\subsection{Modified CAM}
From the finding that On-Off pattern and PathCount contain information of representation, we suggest a type of application, modified Class Activation Maps (CAM) \cite{zhou2016learning}. CAM is a widely used method for interpreting pre-trained DNNs consisting of CONV-FC structure. As the name implies, CAM marks important parts in the input by computing weighted average of convolution activation output using FC weights. We modify Grad-CAM \cite{selvaraju2017grad}, a generalized version of CAM for diverse tasks, by replacing convolution activation with On-Off and PathCount. We call each method, Act-CAM, OnOff-CAM, and PC-CAM. By evaluating the interpretation performance of modified CAMs, we confirm that activation encodes large amount of information in On-Off pattern and PathCount.

We employ two metrics to estimate interpretation performance of modified CAMs, degradation \cite{schulz2020restricting} and tiled-data target matching accuracy that we suggest. For estimating degradation performance, we first quantify importance of each pixel in the input image using interpretability methods such as CAM. Then we evaluate accuracy perturbing the pixels in the order of either most relevant pixels first (MoRF) \cite{samek2016evaluating} or least relevant pixels first (LeRF) \cite{ancona2017towards}. By calculating the area between MoRF and LeRF accuracy trajectories, we measure degradation. If degradation performance of a DNN interpretation method is high, that method tends to figure out important elements of the input well.

Next, we suggest a metric to evaluate the interpretation method when multi-object images are given. We generate multi-object images by tiling four images two by two. As shown in Figure~\ref{fig:tiled-imagenet target-matching accuracy}, we apply interpretation method four times chancing the target class label. Each time, the target class label is inferred based on average importance of each tile. \textit{Target-matching accuracy} is estimated counting the cases where targeted and inferred classes are matched. High accuracy implies that an interpretation method can find out important elements of the input when multi-object images are given.

First, we compare the performance of three modified CAMs using a degradation metric. As CAM is usually employed to find the object related to class decision, we exclude MNIST and CIFAR10 that consist of images where the object fill up the whole image. We present degradation performance for three benchmark models, VGG16, ResNet18, and GoogLeNet in Figure~\ref{fig:CAM}.

In congruence with the On-Off accuracy result in Section 4.2 that On-Off pattern of the last convolution layer effectively represents activation, OnOff-CAM shows degradation performance close to Act-CAM's performance. This result confirms that the class's major information is encoded in the On-Off pattern of the last layer. Meanwhile, PC-CAM's degradation performance is low compared to the other two CAMs.

Second, we provide the target-matching result of modified CAMs. Table~\ref{tab: target-matching accuracy} exhibits that OnOff-CAM scores the best in target-matching accuracy on all three models. The result implies that On-Off is better at pointing out a region related to the target class in multi-object environment than activation.

\begin{table}
\centering
\begin{tabular}{l|ccc}
\toprule
          & \multicolumn{1}{c}{Act-CAM} & \multicolumn{1}{c}{OnOff-CAM} & \multicolumn{1}{c}{PC-CAM} \\ \midrule \midrule
VGG16     & 0.9699                      & \textbf{0.9744}                        & 0.9560                     \\
GoogLeNet & 0.9594                      & \textbf{0.9771}                        & 0.9608                     \\  
ResNet18  & 0.9625                      & \textbf{0.9721}                        & 0.9368                     \\
\bottomrule
\end{tabular}
\newline
\caption{Tiled-Imagenet target-matching accuracy. We present target-matching accuracy of three modified CAMs (Act-CAM, OnOff-CAM, and PC-CAM) using three benchmark models (VGG16, GoogLeNet, and ResNet18) on Tiled-ImageNet data. We generated 5,000 tiled images by randomly selecting concatenating four images. The accuracy is the ratio of correct cases out of 20,000 total target-matching cases.}
\label{tab: target-matching accuracy}
\end{table}


%% file: 5_Conclusion.tex
\section{Conclusion}
To analyze how DNN representation is formed and how it encodes information needed for the task, we devised three experiments with two methodologies, On-Off and PathCount. From the experimental result, we examined that magnitude of representation was not necessarily proportional to PathCount. This result implies that layerwise representation is formed by complex interaction of weights and activation in DNNs. Next, we figured out that replacement of activation with On-Off does not hurt much of the classification performance of pre-trained networks, especially in upper layers. We can conclude that DNN activation relies on On-Off when encoding information for the task. Finally, we provide an application possibility of On-Off pattrn and PathCount by generating modified CAMs, OnOff-CAM and PC-CAM. Even in the target-matching task, OnOff-CAM that is composed of far less information performed better than Act-CAM.